\documentclass[10pt, conference]{IEEEtran}
\IEEEoverridecommandlockouts

\usepackage{amsmath}
\usepackage{amssymb}
\usepackage{algorithm}
\usepackage{algorithmic}
\usepackage{graphicx}
\usepackage{cite}
\usepackage{subfigure}
\usepackage{diagbox}
\usepackage{caption}
\usepackage{color}
\usepackage{bbm}
\usepackage{dsfont}
\usepackage{booktabs}
\usepackage{makecell}
\usepackage[font=small, labelfont=bf, labelsep=period]{caption}
\begin{document}

%\begin{CJK}{UTF8}{gbsn}
\title{\huge Contrastive Language–Image Pre-Training
Model based Semantic Communication Performance Optimization }
% \author{Shaoran Yang\IEEEauthorrefmark{1}, Dongyu Wei\IEEEauthorrefmark{1}, Hanzi Yu\IEEEauthorrefmark{1}, and Mingzhe Chen\IEEEauthorrefmark{1}\IEEEauthorrefmark{2}\\
%  \IEEEauthorrefmark{1}\small Department of Electrical and Computer Engineering, University of Miami, Coral Gables, FL, 33146, USA\\
%  Emails: sxy632@miami.edu, dongyu.wei@miami.edu, hanzhiyu@miami.edu, mingzhe.chen@miami.edu\\
%  }

\author{
    \IEEEauthorblockN{
        Shaoran Yang\IEEEauthorrefmark{1}, 
        Dongyu Wei\IEEEauthorrefmark{1},
        Hanzhi Yu\IEEEauthorrefmark{1},
        Zhaohui Yang\IEEEauthorrefmark{2},
        Yuchen Liu\IEEEauthorrefmark{3},
        and Mingzhe Chen\IEEEauthorrefmark{1}\IEEEauthorrefmark{4}
    }
    \IEEEauthorblockA{
        \IEEEauthorrefmark{1}Department of Electrical and Computer Engineering, University of Miami, Coral Gables, FL, 33146, USA\\
        \IEEEauthorrefmark{2}College of Information Science and Electronic Engineering, Zhejiang University, Hangzhou, China\\
        \IEEEauthorrefmark{3}Department of Computer Science, NC State University, Raleigh, NC, 27695, USA\\
        \IEEEauthorrefmark{4}Frost Institute for Data Science and Computing, University of Miami, Coral Gables, FL, 33146, USA\\
        Emails: \{sxy632, dongyu.wei, hanzhiyu, mingzhe.chen\}@miami.edu, yang\_zhaohui@zju.edu.cn, yuchen.liu@ncsu.edu
        \vspace{-0.5cm}
    }
}

\maketitle
\begin{abstract} In this paper, a novel contrastive language–image pre-training (CLIP) model based semantic
communication framework is designed. Compared to standard neural network (e.g.,
convolutional neural network) based semantic encoders and
decoders that require joint training over a common dataset,
our CLIP model based method does not require any
training procedures thus enabling a transmitter to
extract data meanings of the original data without neural
network model training, and the receiver to train a neural
network for follow-up task implementation without the communications with the transmitter. Next, we investigate the deployment of the CLIP model based
semantic framework over a noisy wireless network. Since the semantic
information generated by the CLIP model is susceptible to
wireless noise and the spectrum used for semantic information transmission is limited, it is necessary to jointly optimize CLIP
model architecture and spectrum resource block (RB) allocation to
maximize semantic communication performance while considering wireless noise, the delay and energy used for semantic communication. To achieve this goal, we use a proximal policy optimization (PPO) based reinforcement learning (RL) algorithm  to learn how wireless noise affect the
semantic communication performance thus finding optimal
CLIP model and RB for each user. Simulation results show that our proposed method improves the convergence rate by up to 40\%, and the accumulated reward by 4x compared to soft actor-critic.
\\
\end{abstract}
\section{Introduction}
With the rapid development of edge devices, such as advanced computing hardware, human intelligence-driven wireless applications with new communication requirements (e.g., data rate, resilience) have emerged. However, traditional communication systems, which primarily focus on bit-level data transmission, are struggling to meet these requirements~\cite{ref1, ref2}. Semantic communications seems a promising solution to meet these emerging application requirements. By leveraging shared knowledge between the transmitter and receiver, semantic communications enables the extraction and transmission of the meaning of data rather than the complete raw data. Thereby significantly enhancing communication efficiency and intelligence~\cite{ref3}. Despite its immense potential, deploying semantic communication over current wireless networks also faces several challenges. Including  efficient semantic information extraction and representation, improving robustness to transmission errors in complex environments, and secure and private semantic communication system design~\cite{ref4}. %Additionally, effectively managing semantic interference in multi-user environments remains a critical issue. 
%\subsection{Related work}

Recently, a number of existing works [5]-[8] has studied the use of deep learning for semantic information extraction and semantic communication performance optimization. In particular, the authors in~\cite{ref5} proposed a semantic communication framework that models the data semantics using knowledge graph and jointly optimizes semantic information extraction and wireless resource management. The authors in~\cite{ref6} used an attention mechanism based reinforcement learning (RL) framework to optimize semantic information extraction and wireless resource allocation strategies. In~\cite{ref7}, the authors designed a lightweighted semantic communication system based on an integrated source and channel coding scheme, and applied model sparsification and quantization to reduce transmission latency.
In~\cite{ref8}, the authors developed a transfer learning-based semantic communication framework for task-unaware and dynamic task request users. However, most of these works require to train the semantic encoder and decoder for a specific user with a target follow-up task (e.g., image regeneration or classification), demands significant time and energy.  

%However, this method lacks consideration for energy consumption and computational complexity.

The main contribution of this work is a novel semantic communication framework that enables 1) a transmitter to extract data meanings of the original data without neural network model training, and 2) the receiver to train a neural network for follow-up task implementation without the communications with the transmitter. In particular, we consider the use of a contrastive
language–image pre-training (CLIP) model as semantic encoder to extract the data meanings of the original data. Compared to standard neural network (e.g., convolutional neural network) based
semantic encoders and decoders that require joint training over
a common dataset, our CLIP model based semantic encoder
does not require any training procedures thus significantly reducing the time and computing power of devices for semantic
encoder deployment. Then, we consider the deployment of CLIP model based semantic encoder and decoder over a large scale network that consists of multiple users and one server. Since the semantic information generated by the CLIP models is susceptible to wireless noise and the spectrum used for semantic information transmission is limited, it is necessary to jointly optimize CLIP model selection and spectrum resource block allocation to maximize semantic communication performance while considering wireless noise, the delay and energy used for semantic communication. This problem is solved by a proximal policy optimization (PPO) based RL algorithm. Simulation results show that our proposed method improves the convergence rate by up to 40\%, and the accumulated reward by 4x compared to soft actor-critic(SAC).

%Simulation results show that our proposed method significantly outperforms the baseline soft actor-critic (SAC) and deep Q-network (DQN) algorithms in terms of both convergence speed and communication efficiency. Compared to SAC, our method achieves a substantially faster convergence rate and a notably higher final accumulated reward. Moreover, our method consistently demonstrates superior performance over DQN across the entire training process. These improvements highlight the effectiveness of the proposed PPO-based optimization strategy in dynamically adapting semantic model selection and spectrum allocation to complex and noisy wireless environments.

%Thus, we have constructed a comprehensive system model that not only encompasses wireless resource allocation based on OFDMA, delay models, and energy consumption models, but also, under strict transmission delay and energy consumption constraints, enables efficient semantic information extraction and transmission through the dynamic selection of an appropriate CLIP model and optimization of resource block allocation. Furthermore, we have employed a reinforcement learning algorithm based on Proximal Policy Optimization (PPO) to dynamically adjust the semantic encoder model selection and resource allocation strategies online, thereby maximizing the performance of subsequent tasks (e.g., image reconstruction or data classification) in complex wireless environments.

\section{System Model}
\begin{figure}[t]
    \centering
    \includegraphics[width=0.45\textwidth]{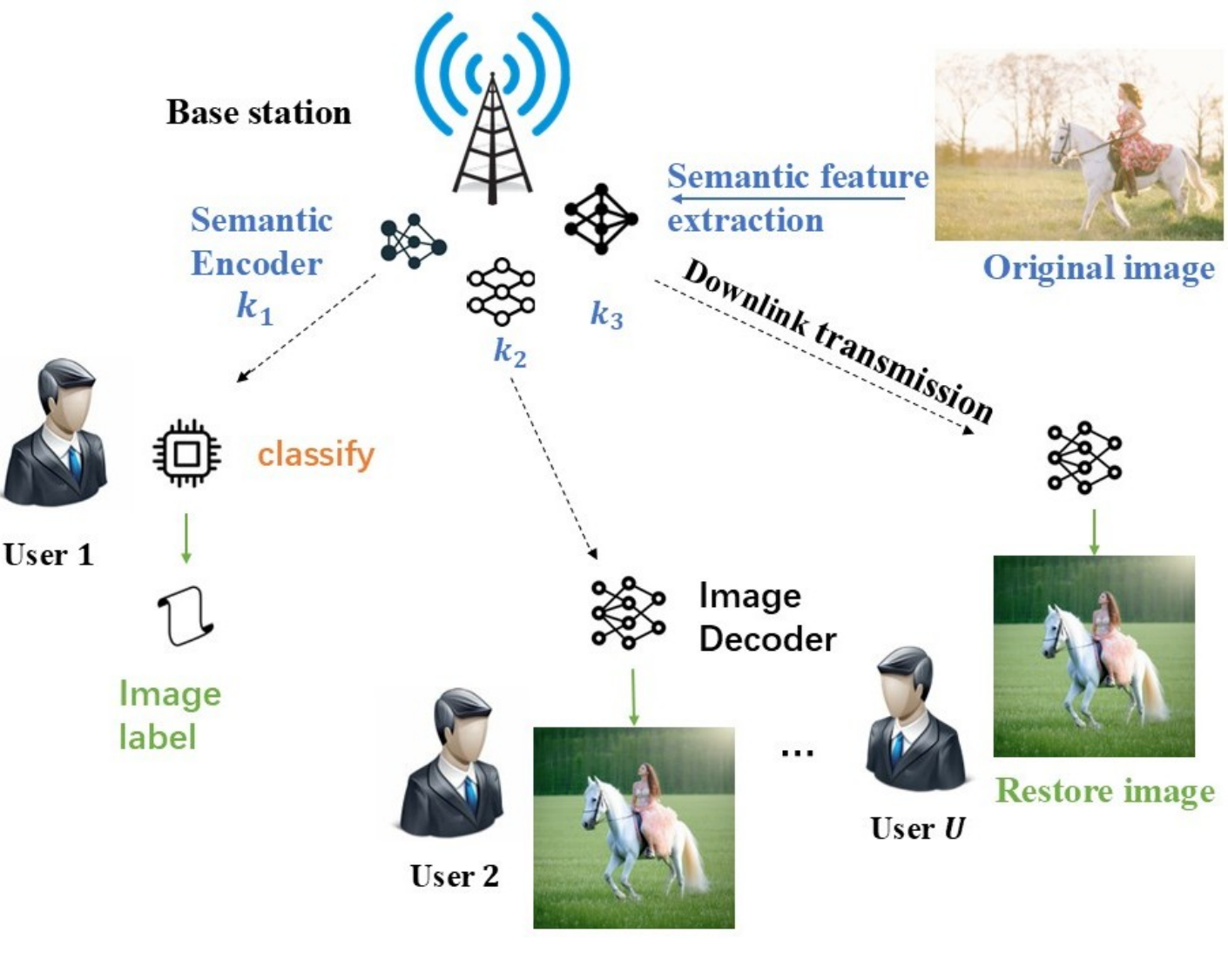} 
    \caption{ Illustration of the proposed semantic communication framework.}
    \label{fig1}
\end{figure}

We consider a wireless network in which a base station (BS) transmits images to a set $\mathcal{U}$ of $U$ users through semantic communication techniques, as shown in Fig. \ref{fig1}. In particular, %The users then reconstruct the complete images based on the received feature vectors. To achieve this image transmission process,
the BS can select an appropriate semantic encoder to extract the feature vectors, called semantic information, from the original images  according to the users' task requirements and wireless channel conditions. The users will use the received semantic information to implement the follow-up tasks (e.g., image regeneration or classification). 
% Specifically, the considered image transmission process consists of four phases: a) using DQN to find the optimal transmission strategy, b) select a variant of a different CLIP model to extract image features and obtain feature vectors based on the optimal strategy, c) distributing data to users according to the optimal strategy, and d) image reconstruction at the user end. 
Next, we first introduce our proposed semantic encoder and decoder. Then, we introduce the transmission and computing model for image processing and transmission. Finally, we explain our considered optimization problem.
\subsection{Semantic Encoder}
Our proposed semantic encoder is based on the CLIP model~\cite{ref9}. Compared to standard neural network (e.g., convolutional neural network) based semantic encoders and decoders that require joint training over a common dataset, our CLIP model based semantic encoder does not require any training procedures thus significantly reducing the time and computing power of devices for semantic encoder deployment. Next, we introduce the components of the CLIP model based semantic encoder. 
\subsubsection{Image feature vector}
The image feature extraction component of the proposed CLIP-based semantic encoder consists of an input layer and a Transformer, as detailed below
% In our research, we selected three different variants of CLIP visual encoders to extract image features. These variants include CLIP-ViT-Large-Patch14, CLIP-ViT-Base-Patch32, and ResNet-101, each demonstrating varying performance levels in the task of image feature extraction.
\begin{itemize}
    \item {Input layer:} Let $\boldsymbol{x_\text{i}}$ be the original image to be sent to the user in the BS. Each image $\boldsymbol{x_\text{i}}$ will be divided into $P$ patches using the patch method \cite{ref13}. Then, each patch is flattened and projected into a $n_\text{e}$-dimensional embedding space with positional embeddings \cite{ref14}. The $n_\text{e}$-dimensional embedding vectors from $P$ patches will be cancated to generate an input vector $\boldsymbol{x}_i'$. 
    \item {Vision transformer:} The vision transformer is used to capture the image features. Here, we use transformer instead of other neural networks since the transformer can use its self-attention mechanism to capture the dependencies between different parts of an image, thus enhancing the understanding of the image. Meanwhile, the transformer allows for the parallel processing of input image data thus improving training and inference speed. 
    The input vector $\boldsymbol{x}_i'$ is fed into a serialized transformer network that consists of $n_\text{l}$ transformer layers. Each transformer layer has $n_\text{h}$ paralleled attention heads. Each attention head $h$ with its unique weights will independently extract features $\boldsymbol{f}_\text{h}$ from the input vector $\boldsymbol{x}_i'$. The feature vectors  $\boldsymbol{f}_\text{h}$ outputted by all attention heads are then concatenated and processed through a feed-forward neural network (FNN). The output of the FNN is an image feature vector which is also the considered semantic information. 
\end{itemize}
\subsubsection{Text fature vector}
The text feature extraction component of the proposed CLIP-based semantic encoder consists of an input layer, a Transformer, and a global text representation, as detailed below:
\begin{itemize}
    \item {Input layer:} %The text feature extraction module of the CLIP model initially processes the input natural language text.
    The original text input by the user is tokenized. Then, each token is mapped to a fixed-dimensional embedding vector through an embedding layer. To preserve the sequential information of the tokens within the text, positional embeddings are added to the word embeddings, forming the final input vector.
    \item {Text transformer:} The generated input vector is fed into the text transformer for text feature extraction. The extraction process is similar to that of the vision transformer, where each layer utilizes parallel attention heads to extract corresponding feature vectors. %After processing through multiple layers, this results in contextually enriched feature vectors for each token. 
    These feature vectors are then concatenated and integrated to generate a text feature vector.
\end{itemize}
\subsection{Image Decoder}
Next, we introduce the use of semantic information (i.e., CLIP model output) for data classification and image regeneration. 

\subsubsection{Image classification} For image classification tasks, we assume that the image that needs to be classified is $\boldsymbol{x}$. Then, the CLIP model based image classification process is summarized as follows:

%first use the CLIP model to generate the CLIP feature vector for the data labels. Then, we can use the CLIP model to generate a CLIP feature vector for the data that needs to be classified. Finally, we compare calculate the cosine similarities between the CLIP feature vector of the data that needs to be classified and  

% For data classification,
% %To meet different usage needs, 
% we use the CLIP model to map a text feature vector and image feature vector of the original data into a shared feature space and calculating their cosine similarities. 
%Classification is achieved by assigning each image to the category whose text feature vector exhibits the highest similarity with the image's feature vector. 
\begin{itemize}
    \item {CLIP model for image label generation:} For each category $v$ of the images, we can generate a text vector ``an image of a {label} $v$" and fed it into the CLIP model to generate the corresponding text feature vector $\boldsymbol{f}_v$. We assume that the dataset has $N$ categories of images. Hence, we will finally generate $N$ CLIP text feature vectors, i.e., $\boldsymbol{f}_1,\ldots, \boldsymbol{f}_{N}$. 
    %These text embeddings were subsequently compared with image feature vectors to perform classification based on the highest similarity scores.
    \item {Similarity calculation:} To classify image $\boldsymbol{x}$, we first use CLIP model to generate its image feature vector $\boldsymbol{f_{x}}$. Then, we calculate the cosine similarity between the image feature vector $\boldsymbol{f_{x}}$ and the text feature vectors $\boldsymbol{f}_1,\ldots, \boldsymbol{f}_N$ of all labels.  
    \item {Image category determination:} We assume that the cosine similarity between $\boldsymbol{f_{x}}$ and $\boldsymbol{f}_{n}$ is $\kappa\bigl(\boldsymbol{f_{x}}, \boldsymbol{f}_{n}\bigr) $. Then, the category of image $\boldsymbol{x}$ is determined by
    \begin{equation}
    \hat{y} = \underset{n \in \{1,\ldots,N\}}{\operatorname{argmax}}\, \kappa\bigl(\boldsymbol{f_{x}}, \boldsymbol{f}_{n}\bigr),
    \end{equation}
    where $\hat{y}$ is the estimated category of the image $\boldsymbol{x}$. 
\end{itemize}
\subsubsection{Image regeneration}
%We have designed a %two image decoder models: 1)
%stable diffusion model based image decoder %and %2) CNN based image decoder. 

% The stable diffusion model based image decoder is used to recover high-resolution images. The CNN based image decoder can use less computational power and regenerate images faster compared to diffusion model based decoder.  -
% \subsubsection{Stable Diffusion based Imgae Decoder}
Our designed image regeneration decoder is based on a stable diffusion model since a stable diffusion model can quickly generate high-resolution images while ensuring that the content accurately reflects user input~\cite{ref10}. 
% Stable Diffusion also demonstrates exceptional stability and robustness, with fewer instances of the common training issues such as mode collapse.
A standard stable diffusion model is primarily used for text-to-image tasks (e.g., using text feature vectors to guide the generation of corresponding images). Here, we investigate the use of image feature vectors $\boldsymbol{f}_\text{h}$ to guide the model to regenerate source images transmitted by the transmitter. 
%Stable Diffusion combines the advantages of Variational Autoencoders (VAEs) and Generative Adversarial Networks (GANs). It generates images that match given conditions from random noise through an iterative process of diffusion and reverse diffusion. 
Our designed stable diffusion model includes an image initialization module, a U-net network \cite{ref11}, and variational autoencoder, which are specified as follows.

\begin{itemize}
    \item{Image initialization module:} The image initialization module is used to generate a latent space representation of an image. This latent space representation matrix serves as the basis for generating source images transmitted by the transmitter. During the training stage, the input of the image initialization module is the source image $\boldsymbol{x}_\text{i}$ and a latent space representation $\boldsymbol{L}$ is generated through the Variational Autoencoder(VAE) encoding module. During the implementation stage, we generate a random Gaussian noise matrix $\boldsymbol{L}$ with dimensions of \(64 \times 64 \times 4\), to approximate the latent space representation of an image. 

    \item {Denoising Diffusion Probabilistic Model(DDPM) scheduler:} The DDPM scheduler controls the forward diffusion of the latent space representation $\boldsymbol{L}$, where noise is incrementally added to the latent space representation $\boldsymbol{L}$ over $T$ time steps.  and generates $T$ noisy latent space representations $\hat{\boldsymbol{L}}_{1}, \ldots, \hat{\boldsymbol{L}}_{T}$. In the reverse diffusion process, DDPM controls how much noise is removed at each step corresponding to the forward process. 
    \item{U-net:} U-Net is used to generate the latent space representations of images by progressively denoising. During the training stage, the input of U-net is are noisy latent space representation $\hat{\boldsymbol{L}}_{t}$ at the current time step, the source image $\boldsymbol{x}_\text{i}$ as label, and the image feature vector (i.e., $\boldsymbol{f}_\text{h}$). Guided by the image feature vector $\boldsymbol{f}_\text{h}$, U-Net denoises the image at the current time step $t$ to generate the noisy latent space representation $\hat{\boldsymbol{L}}_{t-1}$  for the next time step $t-1$. By repeating this process for $T$ iterations, the model progressively removes noise from the image, ultimately obtaining the fully denoised high-quality latent space representation of the image $\boldsymbol{Z}$. During the inference process, the input of the U-net are noisy latent space representation $\hat{\boldsymbol{L}}_{t}$ at the current time step, and the image feature vector (i.e., $\boldsymbol{f}_\text{h}$).
    % It consists of 1) input layer, 2) three CrossAttnDownBlocks and each CrossAttnDownBlock consists of a ResNetBlock and a BasicTransformer, 3)DownBlock, 4)CrossAttnMidBlock module, each composed of ResNetBlock structures, BasicTransformer Block, 5) UpBlock, 6) Three layers of CrossAttnUpBlock module, each consisting of a ResNetBlock structure, a BasicTransformer Block, and an Upsample module, 7) Output Layer composed of GroupNorm, SiLU, and Conv. 
    % Different from standard U-Net, We modified the input layer of the model to accommodate the shape of the image feature vectors and conducted transfer learning based on this adjustment. 
    %The input of U-Net includes $T$ noise latent space representations $\hat{\boldsymbol{L}}_{1},\ldots, \hat{\boldsymbol{L}}_{T}$, and image feature vectors (i.e., $\boldsymbol{f}_\text{h}$) extracted from the CLIP model. The output of U-Net is the latent space representations of images $\boldsymbol{Z}$.
    %For the training process, U-net needs to additionally input the source image $\boldsymbol{x}_\text{i}$ as the label.
    \item{Variational autoencoder decoder:} The VAE decoder is used to transform the output of U-Net to the source image that is transmitted by the transmitter. Hence, the input of VAE decoder is the latent space representations of image $\boldsymbol{Z}$ while the output is the regenerated source image $\hat{\boldsymbol{x}_\text{i}}$.
\end{itemize}

\subsection{Transmission Model}
In our transmission model, %we use Orthogonal Frequency Division Multiple Access (OFDMA) technology to achieve efficient feature vector transmission between the base station and user terminals. Specifically, 
the orthogonal frequency division multiple access (OFDMA) protocol is used for semantic information transmission from transmitter to users. Assume the BS has a set of orthogonal downlink resource blocks (RB) that need to be allocated according to user requirements. For each user $i$ , the RB allocation can be represented by an allocation vector $ \boldsymbol{\alpha}_i = \left[\alpha_{i,1}, \dots, \alpha_{i,q}, \dots, \alpha_{i,Q}\right]$, where $\alpha_{i,q} \in \left\{0, 1\right\} $ with $\alpha_{i,q} = 1$ indicating that RB $q$ is allocated to user $i$, and $\alpha_{i,q} = 0$ , otherwise. 
The transmission data rate of each user $i$ is
\begin{equation}
    c_i\left(\boldsymbol{\alpha}_i\right) = \sum_{q=1}^Q \alpha_{i,q} W \log_2 \left( 1 + \frac{P \phi_i}{I_q + W N_0} \right),
\end{equation} 
where $\alpha_{i,q}$ is the RB allocation index of user $i$, $Q$ is the number of RB, $W$ is the bandwidth of each RB, $P$ is the transmission power of each user, $I_q$ is the interference of RB $q$ caused by the BSs in other service areas, and $N_0$ is the noise power spectral density. The channel gain between the BS and user $i$ is $\phi_i = \gamma_i d_i^{-2}$, where $\gamma_i$ is the Rayleigh fading parameter, and $d_i$ is the distance between the BS and user $i$. 

\subsection{Time Consumption Model}
\subsubsection{Semantic Information Extraction Delay Model}
The time required for the BS to extract the semantic information is
\begin{equation}
    l_{i}^\textrm{B}\left(k_i\right) = \frac{\omega^\textrm{B} D_i^\textrm{X} D_{k_i}^\textrm{M}}{f^\textrm{B}},  \label{eq:semantic information delay}
\end{equation}
where $f^\textrm{B}$ is the frequency of the central processing unit (CPU) clock of each BS, $\omega^\textrm{B}$ is the number of CPU cycles required for computing data (per bit). $ D_i^\textrm{X}$ is the data size of the images $X$ that the BS needs to extract semantic information, $k_i \in \left\{0,1,2\right\}$ is a CLIP model selection index, $D_{k_i}^\textrm{M}$ the size of the CLIP model selected by the BS. 
\subsubsection{Transmission Time}
Given \eqref{eq:semantic information delay}, the time that the BS transmits semantic information to user $i$ is 
\begin{equation}
    l^\textrm{T}_i\left(k_i,\boldsymbol{\alpha}_i\right) = \frac{D_{k_i}^\textrm{O}}{c_i(\boldsymbol{\alpha}_i)}, \label{eq:transmission}
\end{equation}
where $D_{k_i}^\textrm{O}$ is the data size of semantic information (i.e., output of the CLIP model).
\subsubsection{User Computing Model}
Each user needs to use a decoder to recover the original images using the received semantic information. The time required for user $i$ to process this task can be expressed as
\begin{equation}
    l^\textrm{L}_{i}\left(k_i\right) = \frac{\omega^\textrm{U}_{i} D_{k_i}^\textrm{O} D^\textrm{E}}{f^\textrm{U}_{i}}, \label{eq:user computing}
\end{equation}
where $f^\textrm{U}_{i}$  is the frequency of the CPU clock of user $i$, $\omega^\textrm{U}_{i}$ is the number of CPU cycles required for computing the data (per bit) of user $i$, $D^\textrm{E}$ is the size of the decoder model selected by the user $i$. 
\subsubsection{Total time}
Given \eqref{eq:semantic information delay},\eqref{eq:transmission} and \eqref{eq:user computing}, the entire processing time can be expressed as
\begin{equation}
    l_{total}\left(k_i,\boldsymbol{\alpha}_i\right) = l_{i}^\textrm{B}\left(k_i\right) + l^\textrm{T}_i\left(k_i,\boldsymbol{\alpha}_i\right) + l^\textrm{L}_{i}\left(k_i\right),
\end{equation}

\subsection{Energy Consumption Model}
Next, we introduce the energy consumption of the BS extracting and transmitting semantic information, and each user regenerating original images. 
\subsubsection{BS energy consumption}
The energy consumption of the BS extracting semantic information for user $i$ is expressed as
\begin{equation} \label{eq:BS energy}
    e_{i}^\textrm{B}\left(k_i\right) = {\zeta_\textrm{B}\left(f^\textrm{B}\right)^2}  D_i^\textrm{X} D_{k_i}^\textrm{M}+Pl^\textrm{T}_i\left(k_i\right),
\end{equation}
where $\zeta_\textrm{B}$ is the BS energy consumption coefficient. ${\zeta_\textrm{B}\left(f^\textrm{B}\right)^2} D_i^\textrm{X} D_{k_i}^\textrm{M}$ is the energy consumption of the CLIP model extracting semantic information. $Pl^\textrm{T}_i\left(k_i\right)$ is the energy consumption of transmitting semantic information to user $i$.
\subsubsection{User energy consumption}
The energy consumption of user $i$ regenerating original images can be expressed as
\begin{equation}
    e^\textrm{L}_{i}\left(k_i\right) ={\zeta_i\left(f^\textrm{U}_i\right)^2}D_{k_i}^\textrm{O}D^\textrm{E},  \label{eq:user energy}
\end{equation}
where $\zeta_i$ is the user device energy consumption coefficient.

Given \eqref{eq:BS energy} and \eqref{eq:user energy}, the total energy consumption of the BS and user $i$ is
\begin{equation}\label{eq:e}
    e_{total}\left(k_i\right) = e_{i}^\textrm{B}\left(k_i\right) + e^\textrm{L}_{i}\left(k_i\right),
\end{equation}

\subsection{Problem Formulation}
Given the system model, our goal is to optimize the follow-up task performance (e.g., image regeneration or data classification) 
%the mean squared error (MSE) between the raw images and the images regenerated by all users
while meeting the delay and energy consumption requirements. This problem involves the optimization of RB allocation, the selection of the CLIP model based encoder. Let $f\left(\boldsymbol{\alpha}_i,k_i\right)$ be the performance of the follow-up task of user~$i$. In particular, if the follow-up task is image regeneration, $f\left(\boldsymbol{\alpha}_i,k_i\right) $ will be the CLIP Image-to-Image Similarity between the raw images and the images regenerated by the user. If the follow-up task is data classification, $f\left(\boldsymbol{\alpha}_i,k_i\right)$ is the classification accuracy, i.e., $f\left(\boldsymbol{\alpha}_i,k_i\right)=\min_{\boldsymbol{\alpha}_i,k_i} \sum_{i=1}^{U}\sum_{c} \mathds{1}\{y_i = c\}\,\log\left(\hat{y}_i^c\right)$. 
Given these definitions, the optimization problem is formulated as

% If the subsequent task is data classification, the minimization problem can be formulated as
% \begin{equation}
% f\left(\boldsymbol{\alpha}_i,k_i\right)=\min_{\boldsymbol{\alpha}_i,k_i} \sum_{i=1}^{U} \mathbf{1}\{\hat{y}_i \neq y_i\} %\mathcal{L}_{\mathrm{cls}}\left(y_i,\, \hat{y}_i\right) 
% \label{eq:problem}
%\end{equation}
%If the subsequent task is image reconstruction, the minimization problem can be formulated as
\begin{equation}
\min_{\boldsymbol{\alpha}_i,k_i} \sum_{i=1}^{U} f\left(\boldsymbol{\alpha}_i,k_i\right), 
\label{eq:problem}
\end{equation}
%The following constraints need to be met：
\begin{align}
    \text{s.t. } \quad &\alpha_{i,q} \in \left\{0, 1\right\}, \quad k_i \in \left\{0,1,2\right\}, \notag \\
    &\quad \forall i \in \mathcal{U}, \, \forall q \in \mathcal{Q}, \tag{10a} \label{eq:constraint_a}\\
    &\sum_{q=1}^Q \alpha_{i,q} \leq 1, \, \forall i \in \mathcal{U}, \tag{10b} \label{eq:constraint_b}\\
    &\sum_{i=1}^U \alpha_{i,q} \leq 1, \, \forall q \in Q, \tag{10c} \label{eq:constraint_c}\\
    &l_{\text{total}}\left(k_i,\boldsymbol{\alpha}_i\right) \leq D, \tag{10d} \label{eq:constraint_d}\\
    &e_{\text{total}}\left(k_i\right) \leq E, \tag{10e}  \label{eq:constraint_e}
\end{align}
where $\hat{\boldsymbol{x}_i}\left(k_i\right)$ is the image regenerated by user $i$, $\mathcal{Q}$ is the set of RBs that the BS can allocate to the users, $D$ is the maximum delay allowed by the system, $E$ is the maximum energy consumption allowed by the system for each data transmission. 
The constraints from \eqref{eq:constraint_a} to \eqref{eq:constraint_c} guarantee that each user can only occupy one RB and each RB can only be allocated to one user for semantic information transmission. The constraint in \eqref{eq:constraint_d} is a delay requirement of semantic information transmission. The constraint in \eqref{eq:constraint_e} is the energy consumption requirement for semantic communications. 

% Traditional methods have difficulty solving this problem \eqref{eq:problem} for the following reasons. First, the objective function depends not only on the image feature vectors extracted by the BS through the CLIP model, but also on the output of the neural network-based image generation model. Second, the relationship between the mean squared error (MSE) and the optimization variables (i.e., image feature vectors $f_h$ and $\alpha_i$ is challenging to accurately quantify using traditional mathematical approaches. Third, the optimization of this problem depends on the quality of the restored images on the user side. However, in practical scenarios, it is not possible to determine the transmission strategy based on the restoration quality, as the restoration quality can only be assessed after transmission. Therefore, this approach also requires the ability to predict the quality of the restored images.Finally, the function also needs to satisfy the constraint on task time in constraints \eqref{eq:constraint_d} and the constraint on energy consumption in constraints \eqref{eq:constraint_e}.

% To solve this issue, we propose an attention-based reinforcement learning (RL) algorithm that enables the BS to evaluate the corresponding feature vectors based on the quality of image restoration and optimize both resource block allocation and feature vector selection. This approach effectively reduces the total mean squared error (MSE) of the restored images for all users, thereby enhancing image restoration quality.

\section{Proposed Solution}
To solve problem \eqref{eq:problem}, we propose a proximal policy optimization (PPO) based RL algorithm algorithm. 
% In our proposed method, SAC is utilized to optimize the image transmission mechanism between the base station and the users.
%Specifically, SAC determines which CLIP model the base station should use for semantic information extraction, allocates RBs to users for semantc information transmission, and finally determines which model the users should use for image reconstruction. 
Compared to other RL algorithms, PPO has the advantages of high computational efficiency and stable convergence since it improves training robustness by clipping the objective function to prevent overly large policy updates. Next, we will first introduce the components of the proposed PPO algorithm and then explain its training process.

\subsection{Components of the PPO Algorithm}
Our proposed PPO model consists of the following components: 1) agent, 2) state, 3) action, 4) policy, 5) reward function, which are detailed as follows:
\subsubsection{Agent} The agent is the BS which needs to determine the semantic encoder and the RB for each user in order to minimize the objective function in \eqref{eq:problem}.
\subsubsection{State}
The state is used to describe the current network status under which the BS must determine the values of the variables $k_i$ and $\boldsymbol{\alpha}_i $. Hence, a state of the BS includes: 
%1) data requests of the users, which is represented by a vector $\boldsymbol{D}^\textrm{X}=\left[D_1^\textrm{X},\ldots, D_U^\textrm{X} \right]$, 
1) interference over each RB, $\boldsymbol{I}=\left[I_1, \ldots,I_Q\right]$, 2) user location vector $\boldsymbol{p}=\left[\boldsymbol{p}_1, \ldots,\boldsymbol{p}_U\right]$, and 3) available RBs that can be allocated to the users, which is represented by a vector $\boldsymbol{\nu} = \left[\nu_0,\nu_1,\nu_2,...,\nu_Q\right]$ with $\nu_k \in \left\{0,1\right\}$ indicating whether RB $k$ has been allocated to the users with $\nu_k=1$ indicating that RB $k$ has been allocated to the users, otherwise, we have $\nu_k=0$. Given these definitions, each state of the BS at time slot $t$ is $\boldsymbol{s}_t=\left[\boldsymbol{I},\boldsymbol{p},\boldsymbol{\nu} \right]$. %$\boldsymbol{s}_t=\left[\boldsymbol{D}^\textrm{X},\boldsymbol{I},\boldsymbol{p},\boldsymbol{\nu} \right]$.

\subsubsection{Action}
The action of the BS is to determine the CLIP model used by the BS for semantic information extraction, %and the generative model used by users for image reconstruction, 
and the appropriate RBs for image transmission. Hence, at time slot $t$, an action of the BS can be represented by a vector $\boldsymbol{a}_{t}=\left[k_i,  \boldsymbol{\alpha}_i \right]$. Here, at each step, the BS only determines the semantic encoder and RB for only one user. Hence, the BS needs to implement $U$ steps to determine the semantic encoder and RB for all $U$ users.  %From the definition of the action, we see that the BS will determine the semantic encoder and decoder, as well as the RB for only one user at each step.  

% for each user $i \in \left\{1, \dots, U\right\}$. The action can be defined as $\boldsymbol{a}_t = \left[\left(k_1, m_1, \boldsymbol{\alpha}_1\right), \left(k_2, m_2, \boldsymbol{\alpha}_2\right), \dots, \left(k_U, m_U, \boldsymbol{\alpha}_U\right)\right]$.

\subsubsection{Policy}
The policy of the BS is the conditional probability
of the BS choosing action $\boldsymbol{a}_t$ based on state $\boldsymbol{s}_t$.
The policy is approximated by a deep neural network (DNN) parameterized by $\boldsymbol{\varphi}$. The input is the state $\boldsymbol{s}_t$, and the output is the action probability distribution. 
In our problem, the policy network describes the relationship among semantic encoder model selection, RB allocation,  transmission delay, energy consumption, and data transmission quality. The conditional probability of the BS taking action $\boldsymbol{a}_t$ in state $\boldsymbol{s}_t$ is %denoted as 
$\boldsymbol{\pi}_{\boldsymbol{\varphi}}\left(\boldsymbol{a}_t|\boldsymbol{s}_t\right)$.

\subsubsection{Reward}
The reward function $r\left(\boldsymbol{s}_t,\boldsymbol{a}_t\right)$ is used to evaluate the state-action pairs during the entire RL implementation period from time slot $1$ to $U$. %in state $\boldsymbol{s}_t$. Considering our optimization problem \eqref{eq:problem}, once the model selection is determined, the value of MSE for the restored images can be fixed, and the RB only needs to satisfy the constraints \eqref{eq:constraint_a}, \eqref{eq:constraint_b}, \eqref{eq:constraint_c}, \eqref{eq:constraint_d}, and \eqref{eq:constraint_e}. 
Thus, the reward function of the BS at step $t$ is
\begin{equation}
\begin{aligned}
    r\left(\boldsymbol{s}_t,\boldsymbol{a}_t\right) = 
    \begin{cases}
    0,
    &\!\!\!\!\!\! t =1, \ldots, U-1,\\
    %-\sum_{i=1}^U \| \boldsymbol{x}_i - \hat{\boldsymbol{x}_i}\left(k_i\right) \|^2 \\
    %-\sum_{i=1}^{U} \mathbf{1}\{\hat{y}_i \neq y_i\} \\ 
    \sum_{i=1}^{U}\sum_{c} \mathds{1}\{y_i = c\}\,\log\left(\hat{y}_i^c\right) \\
    - \lambda_\textrm{D}  \sum_{i=1}^U \mathds{1}_{\left\{l_\text{total}\left(k_i,\boldsymbol{\alpha}_i\right) > D\right\}} \\
    - \lambda_\textrm{E} \sum_{i=1}^U \mathds{1}_{\left\{e_\text{total}\left(k_i\right) > E\right\}},  
    & t = U,
    \end{cases}
    \label{eq:reward function}
\end{aligned}
\end{equation}
where $\sum_{i=1}^{U}\sum_{c} \mathds{1}\{y_i = c\}\,\log\left(\hat{y}_i^c\right)$ is the sum of the classification cross-entropy loss of all users with $\hat{y}_i^c$ being the estimated probability of data sample belonging to class $c$ and $\mathds{1}_{\left\{\right\}}$ is an indicator function. When $l_\text{total}\left(k_i, \boldsymbol{\alpha}_i\right) > D$, $\mathds{1}_{\left(l_\text{total}\left(k_i, \boldsymbol{\alpha}_i\right) > D\right)} = 1$, otherwise, we have $\mathds{1}_{\left(l_\text{total}\left(k_i, \boldsymbol{\alpha}_i\right) > D\right)} = 0$, %otherwise, $\mathbb{I}_{\left(e_\text{total}\left(k_i, m_i\right) > E\right)} = 1$, 
$\lambda_\textrm{D}$ is the delay penalty coefficient, and $\lambda_\textrm{E}$ is the energy consumption penalty coefficient.

%The reward function constructed in this way can effectively integrate the MSE score with the impacts of delay and energy consumption constraints to find the overall optimal solution.

%\subsubsection{Q function} The Q-function $Q_{\boldsymbol{\theta}}\left(\boldsymbol{s}_t, \boldsymbol{a}_t\right)$, that is approximated by a neural network $\boldsymbol{\theta}$, is used to evaluate the expected reward of the BS taking action $\boldsymbol{a}_t$ in each state $\boldsymbol{s}_t$.
%\subsubsection{Value function}
%The value function $V_{\boldsymbol{\theta}}\left(\boldsymbol{s}_t\right)$, parameterized by $\boldsymbol{\theta}$, is used to estimate the long-term expected reward that BS can obtain starting from each state $\boldsymbol{s}_t$. It is implicitly parameterized by the Q function. The expression for value function is
%\begin{equation}
    %V_{\boldsymbol{\theta}}\left(\boldsymbol{s}_t\right) = \mathbb{E}_{\boldsymbol{a}_t} \left[ Q_{\boldsymbol{\theta}}\left(\boldsymbol{s}_t, \boldsymbol{a}_t\right) - \gamma \log \boldsymbol{\pi}_{\boldsymbol{\varphi}}\left(\boldsymbol{a}_t|\boldsymbol{s}_t\right)\right], \label{eq:value function}
%\end{equation}
%where $\gamma$ is the temperature parameter.
\subsection{PPO Training}
Next, we introduce the training process of the PPO method. 
\subsubsection{Training of the policy neural network}
%In the SAC algorithm, network updates are performed by maintaining three primary loss functions: the loss of the Q network, the loss of the policy network, and (optionally) the loss of the temperature parameter.
The expected reward optimized by the PPO algorithm is
\begin{equation}
    \bar{A}\left(\boldsymbol{\theta}\right) = \mathbb{E}_{ \boldsymbol{a} \sim  \boldsymbol{\pi}_{ \boldsymbol{\theta}}\left( \boldsymbol{s},  \boldsymbol{a}\right)} \left( R\left( \boldsymbol{a}| \boldsymbol{s}\right) \right)
                    \simeq \frac{1}{W} \sum_{w=1}^{W} R\left( \boldsymbol{a}_w^*| \boldsymbol{s}\right) \frac{ \boldsymbol{\pi}_{ \boldsymbol{\theta}}\left( \boldsymbol{s},  \boldsymbol{a}_w^*\right)}{ \boldsymbol{\pi}_{ \boldsymbol{\theta}^*}\left( \boldsymbol{s},  \boldsymbol{a}_w^*\right)} \label{eq:expected reward function}
\end{equation}
To optimize the policy $ \boldsymbol{\pi}_{ \boldsymbol{\theta}}\left( \boldsymbol{s},  \boldsymbol{a}\right)$, we introduce a penalty term to control the difference between the new and old policies
\begin{equation}
    \max_{ \boldsymbol{\theta}} J\left( \boldsymbol{\theta}\right),  \label{eq:policy function}
\end{equation}
where $J\left( \boldsymbol{\theta}\right) = \bar{A}\left( \boldsymbol{\theta}\right) - \lambda f_{KL}\left( \boldsymbol{\pi}_{ \boldsymbol{\theta}^*}\left( \boldsymbol{s},  \boldsymbol{a}\right),  \boldsymbol{\pi}_{ \boldsymbol{\theta}}\left( \boldsymbol{s},  \boldsymbol{a}\right)\right)$ with $\lambda$ being the penalty coefficient, and $f_{KL}\left( \boldsymbol{\pi}_{ \boldsymbol{\theta}^*}\left( \boldsymbol{s},  \boldsymbol{a}\right),  \boldsymbol{\pi}_{ \boldsymbol{\theta}}\left( \boldsymbol{s},  \boldsymbol{a}\right)\right)$ represents the Kullback–Leibler divergence (KLD), which measures the difference between the $\boldsymbol{\pi}_{ \boldsymbol{\theta}^*}\left( \boldsymbol{s},  \boldsymbol{a}\right)$ and $\boldsymbol{\pi}_{\boldsymbol{\theta}}\left( \boldsymbol{s}, \boldsymbol{a}\right)$ policies.
During each iteration $t$, the policy $\boldsymbol{\pi}_{\boldsymbol{\theta}}\left( \boldsymbol{s}, \boldsymbol{a}\right)$ is refined through the standard gradient ascent approach to minimize the total cross-entrop loss. The corresponding policy update rule is given by
\begin{equation}
    \boldsymbol{\theta}^{\left(t\right)} \leftarrow  \boldsymbol{\theta}^{\left(t-1\right)} + \delta \nabla_{\theta} J\left( \boldsymbol{\theta}\right),  \label{eq:update function}
\end{equation}
where $\boldsymbol{\theta}^{\left(t\right)}$ is the parameters of the policy at iteration $t$, $\delta$ is the learning rate. 
\begin{algorithm}[t] \small 
\caption{Training Process of the Proposed PPO Algorithm}
\begin{algorithmic}[1]  
    \STATE \textbf{Input:} Image vector $\boldsymbol{f}_h$ required to transmit to each user, delay threshold $D$, energy consumption threshold $E$, and interference $I_q$ of each RB.
    \STATE \textbf{Initialize:} Parameters $\boldsymbol{\theta}^*$ generated randomly, semantic information extraction model, text recovery model, penalty coefficient $\lambda$, threshold $\tau$, coefficient $\eta$.
    %\STATE Obtain the importance distribution $f(G_i)$ of each semantic information based on (17).
    \REPEAT
        \STATE Store the policy $\boldsymbol{\pi}_{\boldsymbol{\theta}^*}\left(\boldsymbol{s}, \boldsymbol{a}\right)$ and collect $W$ trajectories $\mathcal{W} = \{\boldsymbol{a}_1, \dots, \boldsymbol{a}_W\}$ using $\boldsymbol{\pi}_{\boldsymbol{\theta}^*}\left(\boldsymbol{s}, \boldsymbol{a}\right)$.
        \FOR{$t = 1$ to $T$}
            \STATE Update the parameters of the policy $\boldsymbol{\pi}_{\boldsymbol{\theta}}\left(\boldsymbol{s}, \boldsymbol{a}\right)$ based on \eqref{eq:update function}.
        \ENDFOR
        \STATE Update the penalty coefficient $\lambda$. %based on \eqref{eq: penalty coefficient}.
    \UNTIL{the objective function defined in \eqref{eq:policy function} converges.}
\end{algorithmic} \vspace{-0.1cm}
\end{algorithm}
 
By iteratively updating the policy until the proposed PPO algorithm converges, %the policy parameter $\boldsymbol{\theta}$ can find the relation between the importance distributions of all semantic information and the total MSE. Hence, 
the policy for RB allocation and semantic encoder model selection that minimizes the task training loss can be obtained~\cite{ref15}. The training process of the proposed method is summarized in~\textbf{Algorithm 1}.

\section{Simulation Results and Analysis}
In our simulations, we consider a wireless network with one base station (BS) and $U = 5$ uniformly distributed users. We use the CLIP models in~\cite{ref9} and the stable diffusion model in~\cite{ref2}. The image dataset used to train our proposed PPO algorithm is Open image V6 \cite{ref12}. Other simulation parameters are shown in Table \ref{tb:simu_para}. For comparisons, we compare the proposed method with SAC in \cite{ref16} and  Deep Q-Network(DQN). %We screened some of these categories of images as the training dataset and tested the model's real-world performance. 

\begin{table}[t]\vspace{-0.2cm}
\caption{Simulation Parameters}
\label{tb:simu_para}
\centering
\fontsize{8}{8}\selectfont{
\begin{tabular}{|c||c|c||c|}
\hline
\textbf{Parameters}         & \textbf{Value}  & \textbf{Parameters}         & \textbf{Value}  \\ \hline
$D$ (ms) & $200$ & $E$ (J) & $20$ \\ \hline
$Q$ & 10 & $P$ (W) & $0.2$\\ \hline
$\alpha$ & 2 & $N_0$ (W/MHz
)& $4\times10^{-15}$ \\ \hline
$\lambda_D$ & 1 & $\lambda_E$ & 1 \\ \hline
$W$ (MHz) & 20 & $f^{\textrm{B}}$ (GHz)& $2.5\!-\!3.5$ \\ \hline
$\omega^\textrm{B}
$ (cycles/bit) & $500 \! - \!1000$ & $I_q$ (W) & $10^{-9}\!-10^0$ \\ \hline
%\multicolumn{2}{|c|}{\textbf{optimizer}} & \multicolumn{2}{c|}{\text{our proposed method,\,SAC,\,DQN}} \\ \hline
\multicolumn{2}{|c|}{\textbf{CLIP models}} &
\multicolumn{2}{c|}{%
  \begin{tabular}{@{}c@{}}
    \text{CLIP--ViT--B/32} \\
    \text{CLIP--ViT--B/16} \\
    \text{CLIP--ViT--L/14} 
    \end{tabular}} \\ \hline
\end{tabular}
}
\vspace{-0.5cm}
\end{table}

\begin{figure*}[t]
    \centering
    \includegraphics[width=1\textwidth]{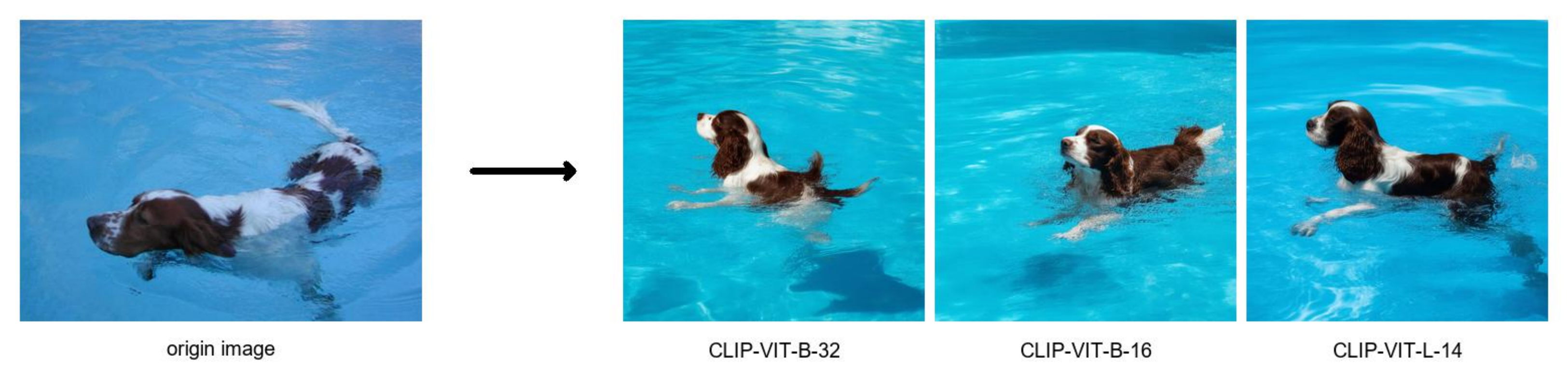} 
    \caption{CLIP models for image regeneration.}
    \label{fig:result1}
    \vspace{-0.5cm}
\end{figure*}

Fig. \ref{fig:result1} shows how the classification accuracy resulting from different CLIP models changes as the channel noise varies. From Fig. \ref{fig:result1}, we see that  the CLIP-ViT-L/14 model achieves higher classification accuracy compared to other two CLIP models. This is because the CLIP-ViT-L/14 model consists of more neural network parameters, thus extracting more features from the original image, and having stronger robustness against noise compared to the other two models. %This allows it to retain more useful information, which helps improve classification accuracy. 
%Meanwhile, due to its larger network capacity, CLIP-ViT-L/14 is able to learn more features of the input images, thus exhibiting stronger semantic modeling capabilities in visual tasks.
%The smaller patch size facilitates the capture of fine-grained visual information, further enhancing the model's ability to distinguish and classify intricate details in images. By incorporating deeper Transformer layers and more attention heads, the model can jointly model both global context and local features, thereby strengthening its representational flexibility. Due to these advantages, the CLIP-ViT-L/14 model achieves superior classification accuracy compared to the other two models.

\begin{figure}[t]
    \centering
    \includegraphics[width=0.45\textwidth]{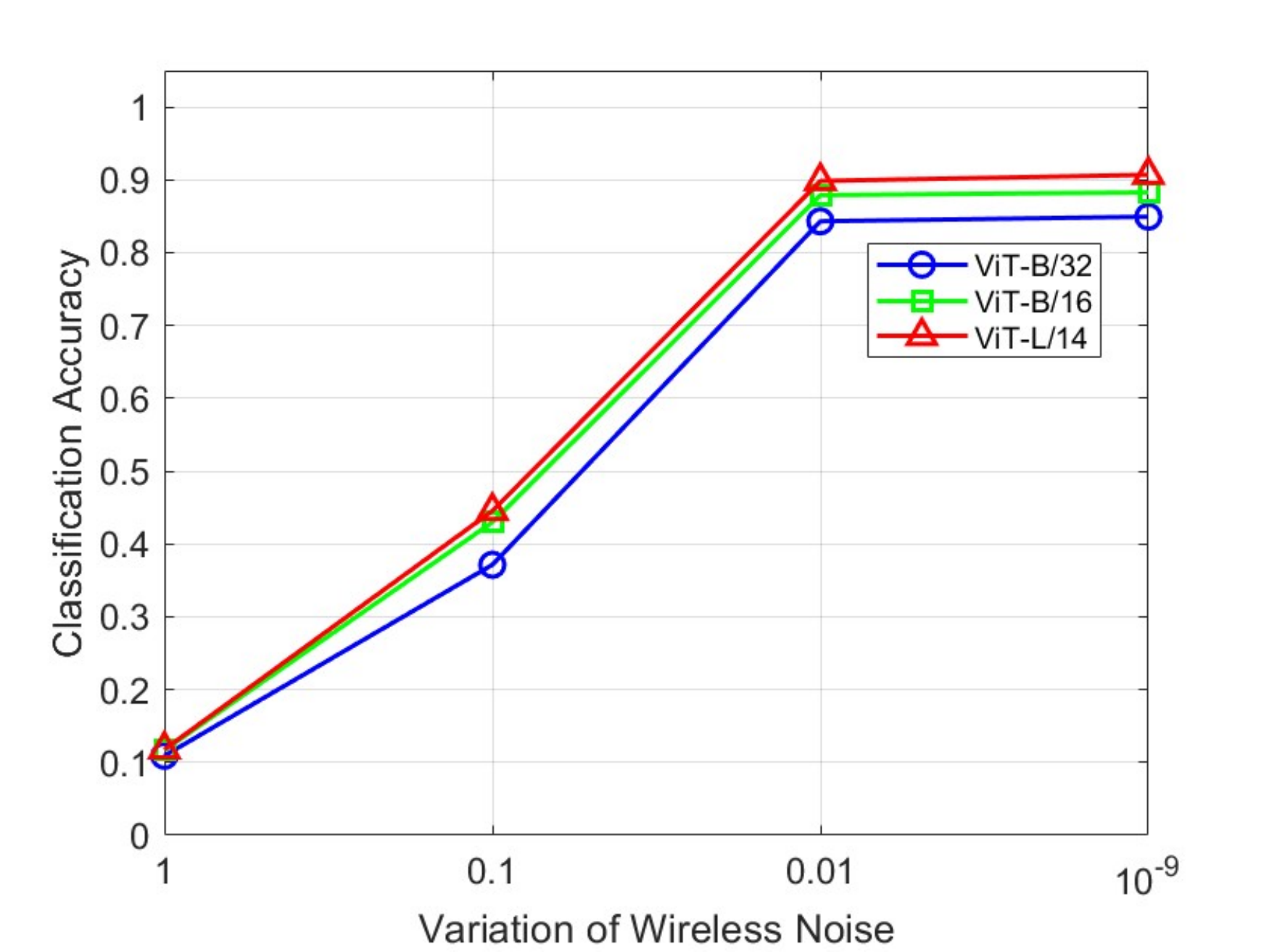} 
    \caption{The classification accuracy of different CLIP models under various noise conditions}
    \label{fig:result2}
    \vspace{-0.5cm}
\end{figure}

Fig. \ref{fig:result2} shows the images generated by our designed stable diffusion model using the text feature vectors extracted by different CLIP models.
%in Fig.2, We compared the images generated by the Stable Diffusion model guided by text feature vectors extracted from different CLIP models. 
From this figure, we see that the image generated based on the semantic features extracted from CLIP-VIT-L/14 has a better reconstruction quality compared to the images generated using the semantic features extracted by the other two CLIP models. This is because CLIP-VIT-L/14 has a stronger model capacity, a larger receptive field, and a better text-vision alignment compared to other two CLIP models. Hence, CLIP-VIT-L/14 can extract more image features to guide stable diffusion for generating higher-quality images.

\begin{figure}[t]
    \centering
    \includegraphics[width=0.45\textwidth]{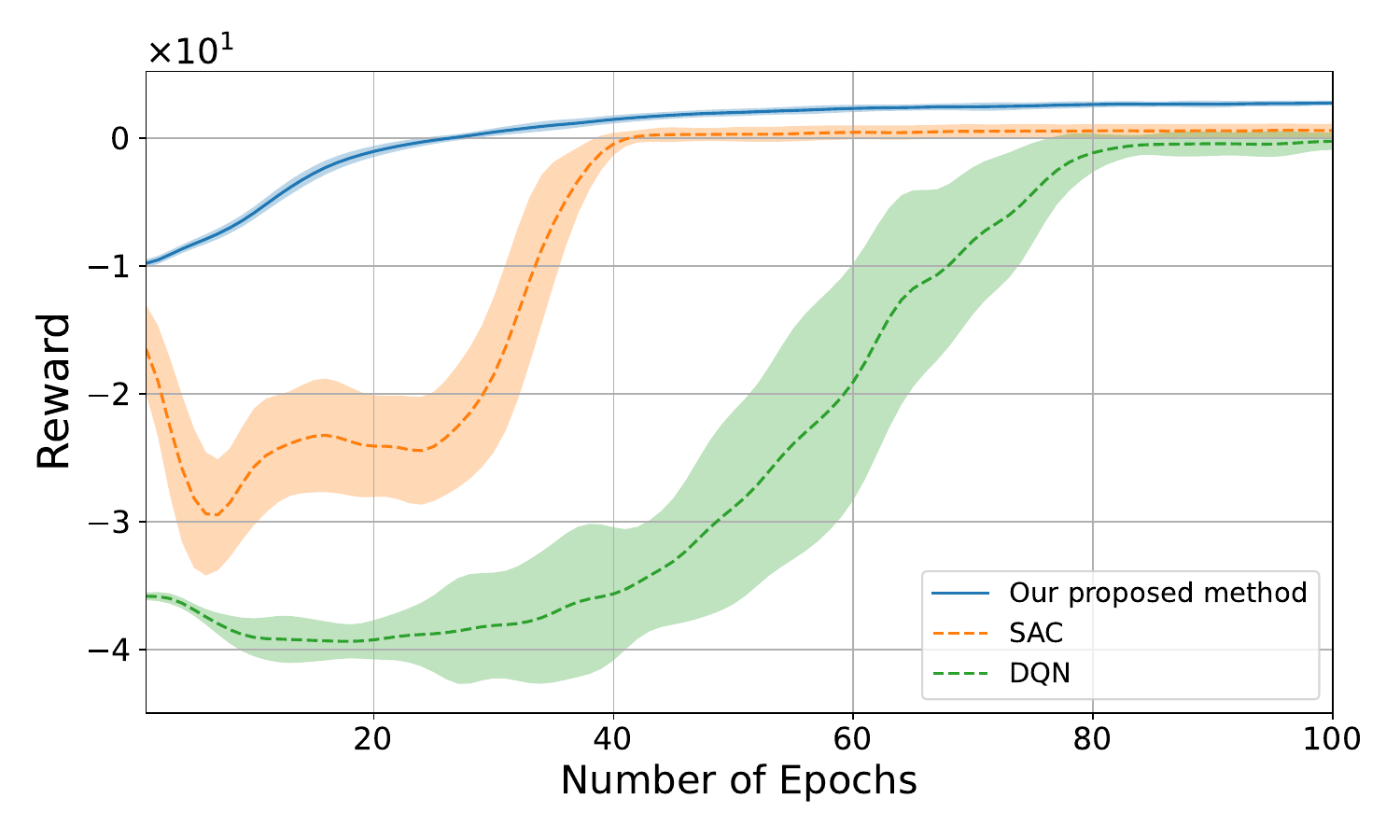} 
    \caption{Convergence of considered algorithms.}
    \label{fig:result3}
    \vspace{-0.6cm}
\end{figure}

Fig. \ref{fig:result3} shows the convergence of considered algorithms during the training process. From this figure, we see that our proposed method can improve the convergence rate by up to 40\%, and the accumulated reward by 4x compared to SAC when the number of epochs is over 100. This is because PPO improves the RL training robustness by clipping the objective function to prevent overly large policy updates.

%constrains policy updates, preventing large shifts that could cause sharp reward drops. %Meanwhile, it handles mixed action spaces without separating them, making it well suited to joint decisions such as selecting a semantic encoder while allocating continuous power or RBs.

\section{Conclusion}

In this paper, we have designed a novel CLIP model based semantic
communication framework is designed. The proposed framework enables a transmitter to
extract data meanings of the original data without neural
network model training, and the receiver to train a neural
network for follow-up task implementation without the communications with the transmitter. Then, we have investigated the deployment of CLIP model based
semantic framework over a large scale network that
consists of multiple users and one server. Since the semantic
information generated by the CLIP model is susceptible to
wireless noise and the spectrum used for semantic information transmission is limited, it is necessary to jointly optimize the CLIP
model architecture and spectrum resource allocation to
maximize semantic communication performance while considering wireless noise, the delay and energy used for semantic communication. To achieve this goal, we have used a PPO-based RL algorithm to learn how wireless noise affect the semantic communication performance thus finding optimal CLIP model and RB for each user. Simulation results show that the designed semantic communication framework yields significant improvements in the performance compared to existing methods.

\def\baselinestretch{0.82}
\bibliographystyle{IEEEbib}
\bibliography{ref}

\begin{thebibliography}{10}

\bibitem{ref1}
X.~Luo, H.-H. Chen, and Q.~Guo,
\newblock ``Semantic communications: Overview, open issues, and future research directions,''
\newblock {\em IEEE Wireless Communications}, vol. 29, no. 1, pp. 210--219, January 2022.

\bibitem{ref2}
Y.~Shi, K.~Yang, T.~Jiang, J.~Zhang, and K.~B. Letaief,
\newblock ``Communication-efficient edge {AI}: Algorithms and systems,''
\newblock {\em IEEE Communications Surveys \& Tutorials}, vol. 22, no. 4, pp. 2167--2191, July 2020.

\bibitem{ref3}
X.~Peng, Z.~Qin, D.~Huang, X.~Tao, J.~Lu, G.~Liu, and C.~Pan,
\newblock ``A robust deep learning enabled semantic communication system for text,''
\newblock in {\em Proc. the IEEE Global Communications Conference}, Rio de Janeiro, Brazil, December 2022, IEEE, pp. 2704--2709.

\bibitem{ref4}
T.~M. Getu, G.~Kaddoum, and M.~Bennis,
\newblock ``Semantic communication: A survey on research landscape, challenges, and future directions,''
\newblock {\em Proceedings of the IEEE}, January 2025.

\bibitem{ref5}
P.~Zhang, W.~Xu, H.~Gao, K.~Niu, X.~Xu, X.~Qin, C.~Yuan, Z.~Qin, H.~Zhao, J.~Wei, et~al.,
\newblock ``Toward wisdom-evolutionary and primitive-concise {6G}: A new paradigm of semantic communication networks,''
\newblock {\em Engineering}, vol. 8, pp. 60--73, January 2022.

\bibitem{ref6}
K.~Niu, J.~Dai, S.~Yao, S.~Wang, Z.~Si, X.~Qin, and P.~Zhang,
\newblock ``A paradigm shift toward semantic communications,''
\newblock {\em IEEE Communications Magazine}, vol. 60, no. 11, pp. 113--119, August 2022.

\bibitem{ref7}
Y.~Wang, M.~Chen, T.~Luo, W.~Saad, D.~Niyato, H.~V. Poor, and S.~Cui,
\newblock ``Performance optimization for semantic communications: An attention-based reinforcement learning approach,''
\newblock {\em IEEE Journal on Selected Areas in Communications}, vol. 40, no. 9, pp. 2598--2613, July 2022.

\bibitem{ref8}
H.~Zhang, S.~Shao, M.~Tao, X.~Bi, and K.~B. Letaief,
\newblock ``Deep learning-enabled semantic communication systems with task-unaware transmitter and dynamic data,''
\newblock {\em IEEE Journal on Selected Areas in Communications}, vol. 41, no. 1, pp. 170--185, November 2022.

\bibitem{ref9}
A.~Radford, J.~W. Kim, C.~Hallacy, A.~Ramesh, G.~Goh, S.~Agarwal, G.~Sastry, A.~Askell, P.~Mishkin, J.~Clark, et~al.,
\newblock ``Learning transferable visual models from natural language supervision,''
\newblock in {\em Proc. International conference on machine learning}. PmLR, July 2021, pp. 8748--8763.

\bibitem{ref13}
A.~Dosovitskiy, L.~Beyer, A.~Kolesnikov, D.~Weissenborn, X.~Zhai, T.~Unterthiner, M.~Dehghani, M.~Minderer, G.~Heigold, S.~Gelly, et~al.,
\newblock ``An image is worth 16x16 words: Transformers for image recognition at scale,''
\newblock {\em arXiv preprint arXiv:2010.11929}, 2020.

\bibitem{ref14}
A.~Vaswani, N.~Shazeer, N.~Parmar, J.~Uszkoreit, L.~Jones, A.~N. Gomez, {\L}.~Kaiser, and I.~Polosukhin,
\newblock ``Attention is all you need,''
\newblock {\em Advances in neural information processing systems}, vol. 30, December 2017.

\bibitem{ref10}
R.~Rombach, A.~Blattmann, D.~Lorenz, P.~Esser, and B.~Ommer,
\newblock ``High-resolution image synthesis with latent diffusion models,''
\newblock in {\em Proc. the IEEE/CVF conference on computer vision and pattern recognition}, New Orleans, Louisiana, USA, June 2022, pp. 10684--10695.

\bibitem{ref11}
Olaf Ronneberger, Philipp Fischer, and Thomas Brox,
\newblock ``U-net: Convolutional networks for biomedical image segmentation,''
\newblock in {\em Proc. Medical Image Computing and Computer-Assisted Intervention}, Munich, Germany, Oct. 2015, Springer, pp. 234--241.

\bibitem{ref15}
J.~Schulman, F.~Wolski, P.~Dhariwal, A.~Radford, and O.~Klimov,
\newblock ``Proximal policy optimization algorithms,''
\newblock {\em arXiv preprint arXiv:1707.06347}, July 2017.

\bibitem{ref12}
A.~Kuznetsova, H.~Rom, N.~Alldrin, J.~Uijlings, I.~Krasin, J.~Pont-Tuset, S.~Kamali, S.~Popov, M.~Malloci, A.~Kolesnikov, et~al.,
\newblock ``The open images dataset v4: Unified image classification, object detection, and visual relationship detection at scale,''
\newblock {\em International journal of computer vision}, vol. 128, no. 7, pp. 1956--1981, March 2020.

\bibitem{ref16}
T.~Haarnoja, A.~Zhou, P.~Abbeel, and S.~Levine,
\newblock ``Soft actor-critic: Off-policy maximum entropy deep reinforcement learning with a stochastic actor,''
\newblock in {\em International conference on machine learning}, Stockholm, Swedem, July 2018, Pmlr, vol.~80, pp. 1861--1870.

\end{thebibliography}
\end{document}